\documentclass[letterpaper, 10 pt, journal, twoside]{ieeetran}
\IEEEoverridecommandlockouts                              

\usepackage{graphics} 
\usepackage{epsfig} 
\usepackage{mathptmx} 
\usepackage{times} 
\usepackage{amsmath} 
\usepackage{amssymb}  
\usepackage{graphicx}
\usepackage{bm}
\usepackage{eucal}

\usepackage{float}
\usepackage[ruled,vlined]{algorithm2e}
\usepackage{pgfplotstable}
\usepackage{pgfplots}
\usepackage{tikz}
\usetikzlibrary[pgfplots.groupplots]
\usepgfplotslibrary{fillbetween}
\pgfplotsset{compat=1.5}
\usepackage{subcaption}
\usepackage{caption}
\usepackage{xcolor}
\usepackage{enumerate}
\usepackage{makecell}
\usepackage{colortbl}

\usepackage{graphicx}
\usepackage{tabularx,ragged2e,booktabs}
\usepackage{url}
\usepackage{siunitx}
\usepackage{rotating}
\usepackage{amssymb}
\usepackage{tabularx}
\usepackage{pgfplots}
\usepackage{csvsimple}
\usepackage{multirow}
\usepackage{cite}
\usepackage{float}
\usepackage{graphicx}
\usepackage{booktabs}
\usepackage{multirow}
\usepackage[font=small]{caption}
\SetKwComment{Comment}{$\triangleright$\ }{}

\SetCommentSty{mycommfont}
\DeclareMathOperator*{\argmax}{arg\,max}

\begin{document}

\title{Active Visuo-Tactile Interactive Robotic Perception \\ for Accurate Object Pose Estimation \\ in Dense Clutter
}

\author {Prajval Kumar Murali, Anirvan Dutta, Michael Gentner, Etienne Burdet, Ravinder Dahiya and Mohsen Kaboli $^{*}$ 
\thanks{Manuscript received: September 9, 2021; Revised January 3, 2022; Accepted January 21, 2022.}
\thanks{This paper was recommended for publication by Editor Dan Popa upon evaluation of the Reviewers' comments.
This work was supported in part by BMW Group and the European Commission via INTUITIVE (Grant agreement ID: 861166) 
} 
\thanks{$^{*}$ Corresponding author: Mohsen Kaboli, mohsen.kaboli@bmwgroup.com  P.K. Murali, A. Dutta, M. Gentner, and M.Kaboli are with the BMW Group, RoboTac Lab, M\"unchen, Germany. 
e-mail: name.surname@bmwgroup.com}%
\thanks{P.K. Murali and R. Dahiya are with the University of Glasgow, Scotland}%
\thanks{A. Dutta and E. Burdet are with The Imperial College of Science, Technology and Medicine, London, England}%
\thanks{M. Kaboli is with the Donders Institute for Brain and Cognition, Radboud University, Netherlands }%
\thanks{Digital Object Identifier (DOI): see top of this page.}
}
\markboth{IEEE Robotics and Automation Letters. Preprint Version. Accepted February, 2022}
{P.K. Murali \MakeLowercase{\textit{et al.}}: Active Visuo-Tactile Interactive Robotic Perception for Accurate Object Pose Estimation in Dense Clutter} 

\maketitle

\begin{abstract}
This work presents a novel active visuo-tactile based framework for robotic systems to accurately estimate pose of objects in dense cluttered environments.
The scene representation is derived using a novel declutter graph (DG) which describes the relationship among objects in the scene for decluttering by leveraging semantic segmentation and grasp affordances networks. The graph formulation allows robots to efficiently declutter the workspace by autonomously selecting the next best object to remove and the optimal action (prehensile or non-prehensile) to perform. Furthermore, we propose a novel translation-invariant Quaternion filter (TIQF) for active vision and active tactile based pose estimation. Both active visual and active tactile points are selected by maximizing the expected information gain. We evaluate our proposed framework on a system with two robots coordinating on randomized scenes of dense cluttered objects and perform ablation studies with static vision and active vision based estimation prior and post decluttering as baselines.
Our proposed active visuo-tactile interactive perception framework shows upto 36\% improvement in pose accuracy compared to the active vision baseline.
\end{abstract}
\begin{IEEEkeywords}
Interactive Perception; Visuo-Tactile Perception; Force and Tactile Sensing; Perception for Grasping and Manipulation
\end{IEEEkeywords}


\section{Introduction}
\label{sec:introduction}
\IEEEPARstart{F}{or} a variety of applications ranging from safe object-robot interaction to robust grasp and manipulation, the ability to accurately estimate the 6 degree-of-freedom (DoF) pose of objects is critical. Especially in unstructured cluttered environments, objects may be occluded from certain viewpoints or may have other objects resting on each other leading to challenging scenarios for accurate object pose estimation.
Such scenarios are common for logistic or retail warehouse robots as well as robots operating inside households.
Interactive perception wherein purposeful physical interactions produce new sensory information to change the state of the environment to enhance perception has been proposed to deal with such scenarios~\cite{kaboli2019tactile,bohg2017interactive, bajcsy1988active}.
Particularly, prehensile and non-prehensile manipulation actions such as grasping or pushing objects can be used to rearrange the cluttered scene to reduce uncertainty in perception~\cite{kaboli2017tactile,kaboliTacManipulation2016, kaboliTacCOM2017}. Such interactive perception maneuvers need to leverage dynamic visual viewpoints as the scene changes upon executing the manipulation actions. 
Furthermore, there might be residual uncertainty in the pose estimate through visual perception due to incorrect calibration of the sensors, environmental conditions (occlusions, variable lighting conditions), or object properties (transparent, specular, reflective)~\cite{kaboli2018robust,Qiang-TRO-2020}. Tactile perception can be used to verify the visual pose estimate to provide a robust and correct pose estimation~\cite{murali2021intelligent,dahiya2019skin, dahiya2019large}.

\begin{figure}[t!]
    \centering
    \includegraphics[width = \columnwidth, height = 5cm]{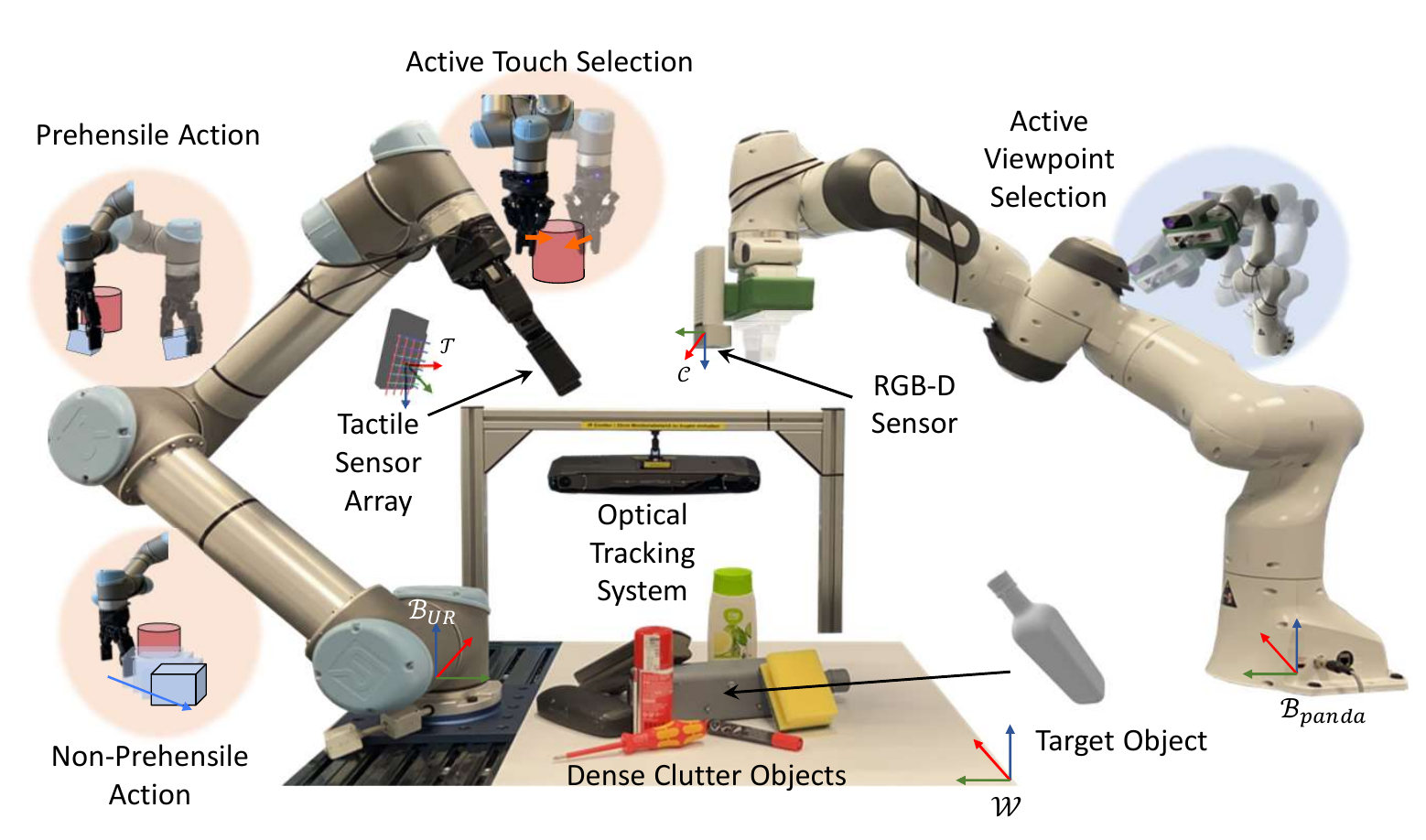}
    \caption{Experimental setup: A Robotiq two-finger adaptive robot
gripper is equipped with 3-axis tactile sensor arrays and mounted on a UR5 robotic arm and a Franka Emika Panda robot with an Azure Kinect (RGB-D) sensor attached to its end-effector. The scene consists of objects placed randomly in dense clutter. An optical tracker is used to provide ground-truth pose of the target object.}
    \label{fig:figure1}
\end{figure}

Vision-based pose estimation in clutter using RGB images or 3D point clouds have been proposed in several works~\cite{mitash2019physics, wu2015active, xiang2018posecnn, potthast2014probabilistic, hinterstoisser2012model}. 
As single viewpoints for pose estimation in clutter is extremely challenging, prior works have used multi-views and combined the observations to recover the object pose~\cite{wu2015active}.
The next best view (NBV) calculation for selecting multiple views have been proposed through information gain metrics such as Shannon entropy~\cite{potthast2014probabilistic}, mutual information~\cite{wang2019efficient} and so on.
As vision-based methods are susceptible to failure in cases of dense object clutter, interactive perception methods have been proposed~\cite{zeng2018learning}. 
Semantic scene understanding methods that are critical for interactive perception have been proposed such as support graphs~\cite{kartmann2018extraction, mojtahedzadeh2015support, schwarz2018fast} which describe the support relationships between objects through geometric reasoning. However, these works abstract the real world objects as simple shapes such as cubes, cylinders and spheres to draw support relations which may not be always applicable in realistic scenes. Similarly, analytical grasp planning relying on geometrical cues may fail in dense clutter with complex objects due to unknown object dynamics. Hence, data-driven approaches have gained popularity for performing manipulation in unstructured scenes~\cite{bohg2013data}.
In~\cite{morrison2020learning}, grasping objects in clutter was demonstrated using generative grasping convolutional neural networks (GG-CNN). 
Zeng et al.~\cite{zeng2018learning} proposed a framework for learning to push and grasp policy that were learnt simultaneously using deep-RL for objects in clutter for grasping applications.
Taking advantage of the synergies of combining prehensile and non-prehensile manipulation actions is of interest while yet to be comprehensively explored by the research community. Furthermore, incorporating mechanisms to choose the best type of action for a given object can increase the autonomy of the robot.

As there may be residual uncertainty with visual estimation, prior works have considered using high-fidelity tactile data to finely localize an object provided with a visual estimate~\cite{bimbo2016hand, piga2021maskukf, murali2021active}.
A known issue in this context is handling the sparsity and density of tactile and visual information respectively. We introduced in~\cite{murali2021active} a novel translation-invariant Quaternion filter (TIQF) for point cloud registration which we extend in this work to active vision-based and active tactile-based pose estimation.
While tactile data can be collected in an uniform or randomized manner or even manually through human tele-operation, these approaches often result in longer data collection time, human intervention and degradation of the sensors due to repeated actions. Hence, active approaches wherein the robot reasons upon the next best action to reduce the collection of redundant data and overall uncertainty of the system have been proposed by Kaboli et. al. in uncluttered scenarios~\cite{kaboli2019tactile, kaboli2018active, kaboli2017tactile, feng2018active}. 
Using their proposed framework, the robotic system autonomously and efficiently explores an unknown workspace to collect tactile data of
the object (construct the tactile point cloud dataset), which
are then clustered to determine the number of objects in the
unknown workspace and estimate the location and orientation of each
object. The robot strategically selects the next position in
the workspace to explore, so that the total variance of the
workspace can be reduced as soon as possible.
Then the robot efficiently learns about the objects’ physical properties, such that with a smaller number of training
data, reliable observation models can be constructed using
Gaussian process for stiffness, surface texture, and center of
mass.

Our contributions are as follows:
\begin{enumerate}[(I)]
    \item A novel graph-based method for autonomous \textit{active} decluttering of the scene, enabling the robot to choose the next object to remove and the optimal action (prehensile or non-prehensile) to perform (Figure~\ref{fig:framework}(a)).
    \item A novel pose estimation method termed \textit{translation-invariant Quaternion filter (TIQF)} for both visual and tactile-based pose estimation.
    \item An active visual viewpoint selection and active tactile touch selection for accurate pose estimation through information gain approach (Figure~\ref{fig:framework}(b)(c)).
    \item Evaluation of the proposed framework on a setup with two robots coordinating to achieve the objective with extensive ablation studies.
\end{enumerate}

\section{Methods}
\label{sec:methods}
\subsection{Problem Formulation and Proposed Framework}
\begin{figure*}[ht!]
    \centering
    \includegraphics[width = 0.9\textwidth, height = 9cm]{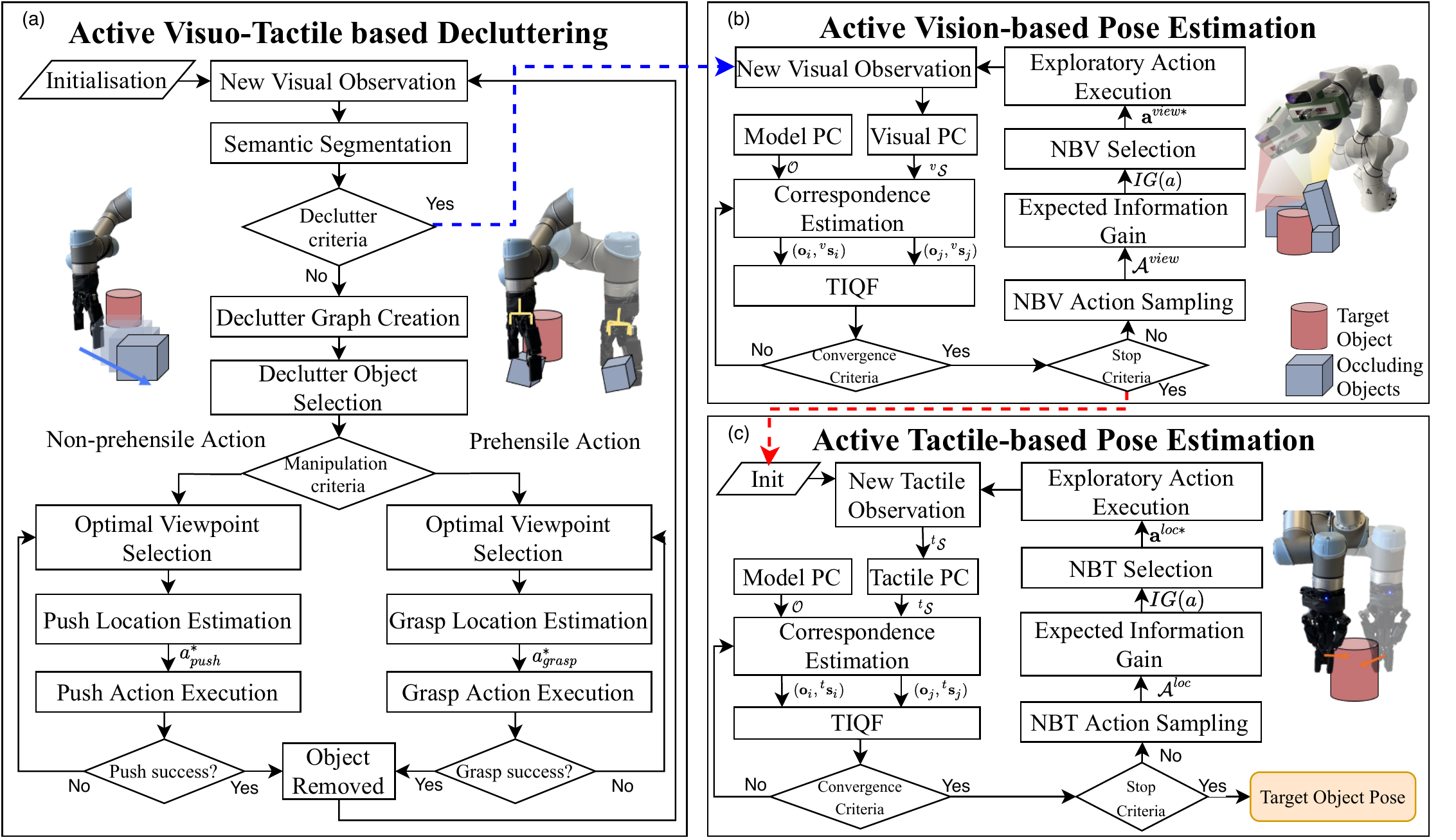}
    \caption{The proposed framework for active visuo-tactile interactive perception for object pose estimation in dense clutter.}
    \label{fig:framework}
\end{figure*}
We propose a novel framework shown in Figure~\ref{fig:framework} to robustly estimate the 6 DoF pose of a known object of interest or target object through active visuo-tactile perception in dense clutter by interactively decluttering the other objects in the workspace. 
Firstly, the robot deterministically declutters the workspace by using either prehensile or non-prehensile actions. This provides the flexibility to choose the action with the highest probability of success. The robot reasons upon the next object to remove to declutter the workspace with minimal actions. Secondly, upon sufficient decluttering the robot actively chooses viewpoints for vision-based pose estimation using an information gain approach. 
Finally, an active tactile based pose estimation is performed to \textit{correct and verify} the visual pose estimate. 

\subsection{Active Decluttering of the Workspace}
\label{sec:declutter}
In order to interactively declutter the workspace, the physical geometry relations between various objects in clutter are autonomously inferred. We define a directed scene graph in form of a tree termed \textit{declutter graph} $\mathcal{G} = (\mathcal{V}, \mathcal{E})$ wherein the vertices in $\mathcal{V}$ represent the various objects $O_i$ in the scene and the edges $\mathcal{E}$ define the action to be used to declutter the object. 
The root node of the graph $\mathcal{G}$ is the target object $O_T$, which we seek to localize. Furthermore, the graph explicitly encodes the next object to be removed by computing a weight signifying how much it occludes $O_T$ and the associated action (grasp or push). The steps in building the graph are depicted in Figure~\ref{fig:declutter_graph}(a). From a cluttered scene, a RGB image and a depth image are taken as inputs to our framework. We use a state-of-the-art semantic segmentation network~\cite{chen2017deeplab} and grasp affordance network~\cite{morrison2020learning} on the RGB image and depth image to extract the semantic segmentation $\mathcal{M}_{seg}$ and grasp success metrics $q_k \in [0,1]$ respectively. We adapted the pretrained segmentation network~\cite{chen2017deeplab} with our own dataset consisting of different objects in clutter and their respective segmentation masks. 

For two objects $O_i, O_j$ an edge $e_{ij}$ $\in$ $\mathcal{E}$ is added if the overlap-metric is above a threshold $\mu_o$ or the minimum distance between the contours $d_{ij}$ is below $\mu_d$. Thus, an edge $e_{ij}$ $\in$ $\mathcal{E}$ is given by
\begin{equation}
        e_{ij} = 
                \begin{cases}
                  IoU_{ij}   &\text{if }  (IoU_{ij} > \mu_{o}) \\ 
                  1/d_{ij}    &\text{if }  (d_{ij} < \mu_{d}) \wedge (IoU_{ij} \leq \mu_{o}) \\
                  0  &\text{otherwise} .
                \end{cases}
    \label{eq:edge}
\end{equation}
The \textit{Intersection Over Union (IoU)} is used as overlap measure with $IoU_{ij} = \mathcal{C}_i \cap \mathcal{C}_j / \mathcal{C}_i \cup \mathcal{C}_j$, where $\mathcal{C}$ defines all points belonging to the minimum area bounding box of the contours of the respective object masks. The threshold values $\mu_o$ has been tuned empirically to be 0.05 and $\mu_d$ to 0.5. 
Subsequently, an action attribute is added to each edge of the graph. Starting with the leaf vertices, for each vertex $O_k$, we attribute the incoming edge to the vertex, $e_{ik} \in \mathcal{E}$ in the graph with a prehensile or non-prehensile action $a_k$ to declutter according to a grasp quality value $q_k$ as: \begin{gather}
        a_k = 
                \begin{cases}
                  a_k^{grasp}   &q_k \geq \mu_{q}\\
                  a_k^{push}   &q_k < \mu_{q}
                \end{cases}
              .
    \label{eq:grasppush}
\end{gather}
We use a mix of both types of actions as for some objects with peculiar shapes, it is challenging for the robot to perform prehensile grasp actions whereas push may be simpler. Since the goal is to declutter the scene in a deterministic manner, Equation~\ref{eq:grasppush} ensures that if an object can be grasped with high confidence, a grasp action is executed. If the confidence is below the threshold $\mu_{q}$, a push action is executed.
The value of $\mu_q$ has been empirically set to 0.1.
The object to be removed next is inferred from the leaf nodes with the highest valued ${e}_{ik}$ defined in Equation~\ref{eq:edge}.

\subsubsection{Push Action} 
\label{sec:push_action}
\begin{figure}[ht!]
    \centering
    \includegraphics[width = \columnwidth, height = 7.5cm]{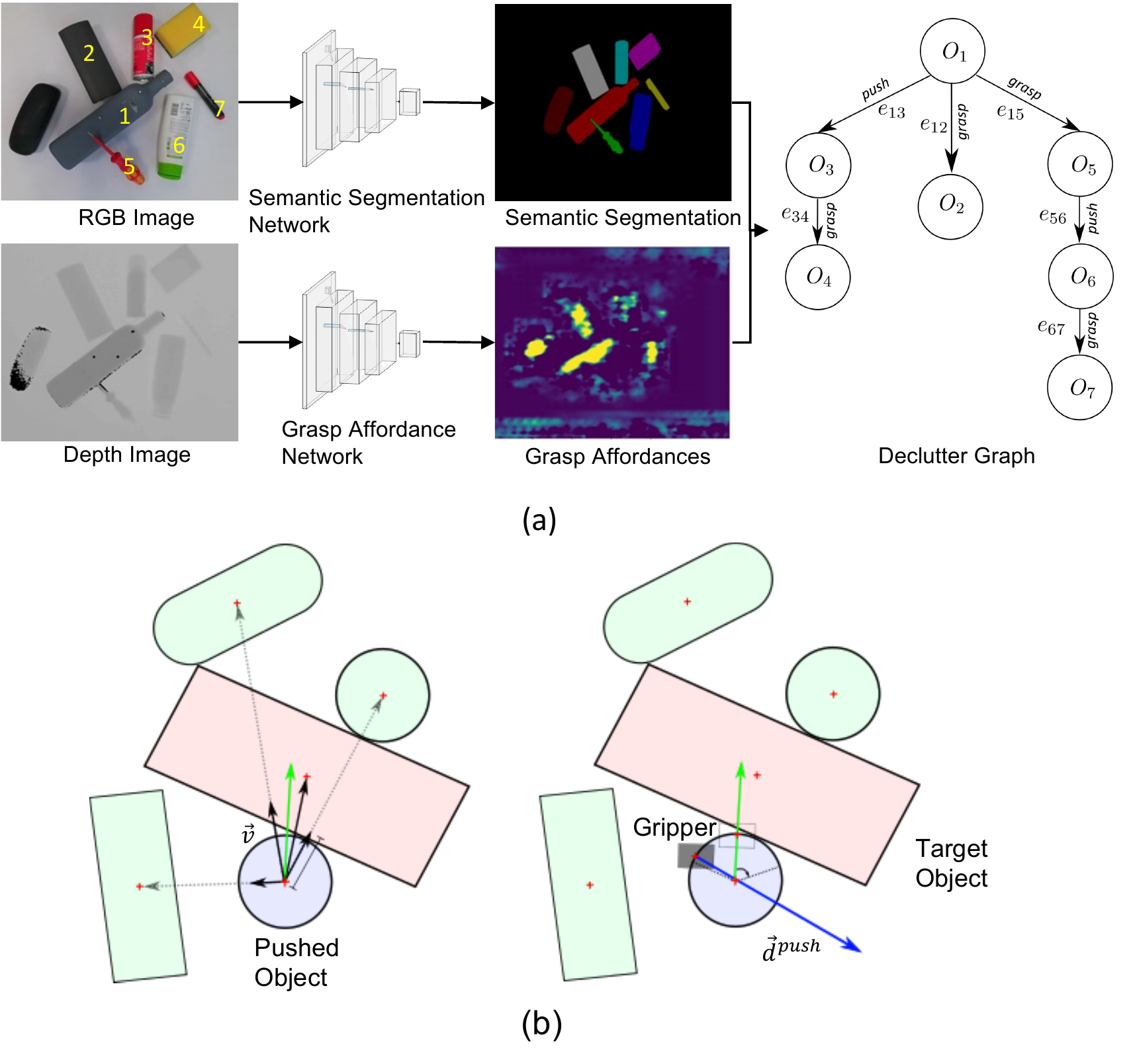}
    \caption{(a) Pipeline for the declutter graph from the semantic segmentation network and the grasp affordance network. (b) Push action formulation}
    \label{fig:declutter_graph}
\end{figure}

We parameterize the push action by a tuple composed of a push point and direction, i.e., $a^{push} = (\mathbf{p}^{push}, \overrightarrow{\mathbf{d}}^{push})$. The trajectory of pushing is a straight line for a fixed predefined distance. We further assume \textit{quasi-static pushing}~\cite{mason1986scope} and that the object moves on a flat 2D surface. 
Given the segmentation mask, we compute vectors $\mathbf{v}_{i,k} \forall i$ between the centroid of the bounding box of each object and the object to be pushed. The vector pointing towards the clutter is then given by $\mathbf{v} = \sum_{i} w_i\mathbf{v}_{i,k}$. Therein each vector is weighted with $w_i$, such that objects that are further away, have less influence on the direction. Finally, the push direction is obtained from $\overrightarrow{\mathbf{d}}^{push} = -\frac{\mathbf{v}}{|| \mathbf{v}||}$. The push point $p^{push}$ is calculated as the point at the intersection of the contour of the segmentation mask and push direction $\overrightarrow{\mathbf{d}}^{push}$ placed at the centroid, as shown in Figure~\ref{fig:declutter_graph}(b). This ensures the push action is aligned towards the centroid of the object. However, due to the width of the fingertips of the gripper, it is not always possible to reach this point due to surrounding clutter. To incorporate this constraint, we sample points in the vicinity of the touch point, place a bounding box in the size of the gripper and calculate the mean \textit{IoU} with all objects. The size of the gripper is calculated by projecting the real world gripper length in the image plane using the transformation between the world frame $\mathcal{W}$ and camera frame $\mathcal{C}$ at the configuration where the push affordance is computed. The point leading to the smallest mean \textit{IoU} is chosen as $\mathbf{p}^{push}$. Furthermore, we use the tactile sensors embedded in the gripper to detect a loss of contact during push which stops the execution and triggers a recalculation of the push action.
\subsubsection{Grasp Action}
\label{sec:grasp_action}
We used the generative grasping CNN (GG-CNN)~\cite{morrison2020learning} for providing grasp affordances in terms of the grasp position, the orientation and the probability of success of the grasp given by the quality measure $q_k$, which is also used in the declutter graph creation. Since we require object specific grasping, we use our semantic segmentation output to mask the depth image input to GG-CNN. Furthermore, in order to improve the object specific grasp estimates, we move the robot to a new viewpoint above the centroid of the chosen object given by the segmentation mask at a predefined height. The grasp action $a^{grasp}$ is defined by a grasp point $\mathbf{p}^{grasp}$, a grasp angle $\alpha^{grasp}$ and an end point to place the object at $\mathbf{p}^{place}$ as a tuple $(\mathbf{p}^{grasp}, \alpha^{grasp}, \mathbf{p}^{place})$. If the grasp action fails during execution detected by a loss of contact using tactile sensors, the execution is stopped and recalculation of the grasp action is triggered using the vision sensor.

\subsection{Translation-Invariant Quaternion Filter (TIQF) for Pose Estimation}
\label{ssec:tiqf}
We tackle the \textit{active} visual and \textit{active} tactile pose estimation problem via a Bayesian-filter based approach termed as translation-invariant quaternion filter (TIQF). The TIQF is a sequential filtering method for point cloud registration that is applicable to sparse as well as dense point clouds. 
Point cloud registration problem given known correspondences can be formalised as
\begin{equation}
     \mathbf{s}_i = \mathbf{R}\mathbf{o}_i + \mathbf{t} \quad i = 1, \dots N,
     \label{eq:generativemodel}
 \end{equation}
where $\mathbf{s}_i \in \mathbb{R}^3$ are points in the scene cloud $\mathcal{S}$ extracted from sensor measurements and $\mathbf{o}_i \in \mathbb{R}^3$ are the corresponding points belonging to the model cloud $\mathcal{O}$ extracted from the model mesh, $\mathbf{R} \in SO(3)$ and $\mathbf{t} \in \mathbb{R}^3$ are the unknown rotation and translation respectively which aligns $\mathbf{o}_i$ to $\mathbf{s}_i$.
We decouple the rotation and translation estimation by finding the relative vectors between a pair of corresponding points as $\mathbf{s}_{ji} = \mathbf{s}_j - \mathbf{s}_i$ and $\mathbf{o}_{ji} = \mathbf{o}_j - \mathbf{o}_i$. This simplifies Equation~\eqref{eq:generativemodel} as:
\begin{align}
    \mathbf{s}_j - \mathbf{s}_i &= (\mathbf{R}\mathbf{o}_j + \mathbf{t}) - (\mathbf{R}\mathbf{o}_i + \mathbf{t}) ,\\
    \mathbf{s}_{ji} &= \mathbf{R}\mathbf{o}_{ji}.
    \label{eq:trans_invariance}
\end{align}
As Equation~\eqref{eq:trans_invariance} is independent of $\mathbf{t}$, this is termed as \textit{translation-invariant measurements}. Given a rotation estimate $\hat{\mathbf{R}}$, the translation estimate $\hat{\mathbf{t}}$ can be found in closed form solution as:
\begin{equation}
    \hat{\mathbf{t}} = \frac{1}{N} \sum_{i=0}^{N} (\mathbf{s}_i - \hat{\mathbf{R}} \mathbf{o}_i).
    \label{eq:translation_solution}
\end{equation}
To estimate rotation, we cast the problem into a Bayesian estimation framework. We denote the rotation estimate $\hat{R}$ in its quaternion form as the state $\mathbf{x}$ which needs to be identified through measurements $\mathbf{z}$ obtained via actions $\mathbf{a}$ upto time $t$. Upon decluttering, the objects' pose remain unaltered during active vision-based and tactile-based pose estimation as we perform guarded touch actions~\cite{petrovskaya2016active}. Hence the state estimate is provided by a recursive Bayes filter as: 
\begin{align}
    p(\mathbf{x} | \mathbf{z}_{1:t}, \mathbf{a}_{1:t}) 
    &= \eta p(\mathbf{z}_{t} | \mathbf{x}, \mathbf{a}_{t}) p(\mathbf{x}| \mathbf{z}_{1:t-1}, \mathbf{a}_{1:t-1}),
    \label{eq:bayesian_filter:1}
\end{align}
where $\eta$ is a normalization constant. We estimate the current belief $p(\mathbf{x} | \mathbf{z}_{1:t}, \mathbf{a}_{1:t})$ through a Kalman filter. To derive a  linear filter, we derive a linear state and measurement model. We reformulate Equation~\eqref{eq:trans_invariance} using quaternions as:
\begin{equation}
    \widetilde{\mathbf{s}}_{ji} = \mathbf{x} \odot \widetilde{\mathbf{o}}_{ji} \odot \mathbf{x}^{*}, 
    \label{eq:quat_objective}
\end{equation}
where $\odot$ is the quaternion product, ${\mathbf{x}}^{*}$ is the conjugate of $\mathbf{x}$, and $\widetilde{\mathbf{s}}_{ji}=\{0,\mathbf{s}_{ji}\}$ and $\widetilde{\mathbf{o}}_{ji}=\{0,\mathbf{o}_{ji}\}$.
As $\mathbf{x}$ is an unit quaternion, using the fact that $\mathbf{x}^{*}\odot \mathbf{x} = \mathbf{x}\odot \mathbf{x}^{*} = 1$ to get:
\begin{align}   
    \widetilde{\mathbf{s}}_{ji}\odot \mathbf{x} &- \mathbf{x} \odot \widetilde{\mathbf{o}}_{ji} = \mathbf{0}.
    \label{eq:quat_objective_2}
\end{align}
We can rewrite Equation~\eqref{eq:quat_objective_2} as:
\begin{align}
    \begin{bmatrix}
        0 & -\mathbf{s}_{ji}^T \\
        \mathbf{s}_{ji} & \mathbf{s}_{ji}^{\times}
    \end{bmatrix}\mathbf{x} -  \begin{bmatrix}
        0 & -\mathbf{o}_{ji}^T \\
        \mathbf{o}_{ji} & -\mathbf{o}_{ji}^{\times}
    \end{bmatrix} \mathbf{x} = \mathbf{0} \\
    \begin{bmatrix}
        0 & -(\mathbf{s}_{ji} - \mathbf{o}_{ij})^T \\
        (\mathbf{s}_{ji} - \mathbf{o}_{ji}) & (\mathbf{s}_j + \mathbf{s}_i + \mathbf{o}_j + \mathbf{o}_i)^{\times}
        \end{bmatrix}_{4 \times 4} \mathbf{x} &= \mathbf{0} 
        \label{eq:expected_measurement}
\end{align}
where $[\ ]^\times$  is the skew-symmetric matrix form.
Equation~\eqref{eq:expected_measurement} is of the form $\mathbf{H}_t\mathbf{x} = 0$ where $\mathbf{H}_t$ is the \textit{pseudo-measurement} matrix such that
\begin{align}
\mathbf{H}_t
=     \begin{bmatrix}
        0 & -(\mathbf{s}_{ji} - \mathbf{o}_{ji})^T \\
        (\mathbf{s}_{ji} - \mathbf{o}_{ji}) & (\mathbf{s}_j + \mathbf{s}_i + \mathbf{o}_j + \mathbf{o}_i)^{\times}
        \end{bmatrix}_{4 \times 4} 
        \label{eq:pseduo_meas_mat}
\end{align}
The Equation~\eqref{eq:expected_measurement} represents a noise-free state estimation where $\mathbf{H}_t$ solely depends on the corresponding measurements.
It can be inferred that $\mathbf{x}$ must lie in the nullspace of $\mathbf{H}_t$.
Similar to~\cite{srivatsan2016estimating}, we design a pseudo-measurement model as:
\begin{align}
    \mathbf{H}_t \mathbf{x} &= \mathbf{z}^h,
    \label{eq:measurement_model}
\end{align}
wherein we enforce the pseudo-measurements $\mathbf{z}^h = 0$. 
As we assume the $\mathbf{x}$ and $\mathbf{z}_t$ to be Gaussian distributed and a static process model, the resulting Kalman equations are given by:
\begin{align}
    \mathbf{x}_{t} &= \bar{\mathbf{x}}_{t-1} - \mathbf{K}_t \left( \mathbf{H}_t \bar{\mathbf{x}}_{t-1} \right) \\
    \Sigma^{\mathbf{x}}_{t} &= \left( \mathbf{I} - \mathbf{K}_t \mathbf{H}_t \right) \bar{\Sigma}^{\mathbf{x}}_{t-1} \\
    \mathbf{K}_t &= \bar{\Sigma}^\mathbf{x}_{t-1} \mathbf{H}_t^T \left( \mathbf{H}_t\bar{\Sigma}^\mathbf{x}_{t-1} \mathbf{H}_t^T + \Sigma_t^{\mathbf{h}}\right)^{-1}, 
    \label{eq:kalman_equations}
\end{align}
where $\bar{\mathbf{x}}_{t-1}$ is the normalized mean of the state estimate at $t-1$, $\mathbf{K}_t$ is the Kalman gain and $\bar{\Sigma}^{\mathbf{x}}_{t-1}$ is the covariance matrix of the state at $t-1$. 
The parameter $\Sigma_t^{\mathbf{h}}$ is the measurement uncertainty at timestep $t$ which is state-dependent and is defined as follows~\cite{choukroun2006novel}:
\begin{align}
    \Sigma_t^{\mathbf{h}} = \frac{1}{4}\rho\left[ tr(\bar{\mathbf{x}}_{t-1}\bar{\mathbf{x}}_{t-1}^T + \bar{\Sigma}^{x}_{t-1})\mathbb{I}_4 - (\bar{\mathbf{x}}_{t-1}\bar{\mathbf{x}}_{t-1}^T + \bar{\Sigma}^{x}_{t-1} )\right], 
    \label{eq:choukron}
\end{align}
where $\rho$ is a constant which corresponds to the uncertainty of the correspondence measurements and is empirically set to 0.05 and $tr$ refers to trace.
However, Kalman filter does not preserve the constraints on the state-variables such as the unit-norm property of the quaternion in our case~\cite{choukroun2006novel}. Hence, a common technique is to normalise the state and the associated uncertainty after each update:
\begin{align}
    \bar{\mathbf{x}}_{t} = \frac{\mathbf{x}_{t}}{||\mathbf{x}_{t}||_2} \quad \bar{\Sigma}^{\mathbf{x}}_{t} = \frac{\Sigma^{\mathbf{x}}_{t}}{||\mathbf{x}_{t}||_2^2}.
\end{align}
The rotation estimate $\bar{\mathbf{x}}$ (quaternion) is converted to $\mathbf{R} \in SO(3)$ and used to estimate the translation according to Equation~\eqref{eq:translation_solution}.
Thus, with each iteration we obtain a new rotation and translation estimate which is used to transform the model. The transformed model is used to recompute correspondences and repeat the Kalman Filter update steps. We calculate the change in homogeneous transformation between iterations $\Delta_{TIQF} < \xi^{conv}$ i.e., if the difference in the output pose is less than a specified threshold which in our experiments is $0.1mm$ and $0.1^o$ respectively and/or maximum number of iterations in order to check for convergence ($max\_it_{TIQF} = 100$).

\subsection{Next Best Action for Pose Estimation}
\label{ssec:nbv}
\subsubsection{Next Best View (NBV) Selection}
The next best view (NBV) problem seeks to find the most optimal next view point to observe an environment given previous measurements by minimising some aspect of the unobserved space through an objective function~\cite{potthast2014probabilistic}.
In comparison to existing approaches for NBV which is used for mapping the entire environment~\cite{potthast2014probabilistic} or for object reconstruction~\cite{delmerico2018comparison}, we design an object-driven active exploration method for object pose estimation.

We extract the approximate centroid of the current target object from our semantic segmentation network. We capture a point cloud from an initial view that is randomly sampled within the constraints of the workspace and the robot. The semantic segmentation output is used to crop the entire point cloud around the region of interest of the target object. We discretize the resulting point cloud into a 3D occupancy grid $\mathcal{OG}$ with resolution $g_{res}$.
Each cell $c_i$  in the occupancy grid is represented by a Bernoulli random variable and has an occupancy probability $p(c_i)$. There are two possible states for each cell with $c_i = 1$ indicating the cell is occupied and $c_i = 0$ for an empty cell. 
A common independence assumption of each cell with other cells enables the calculation of the overall entropy of the occupancy grid as the summation of the entropy of each cell. 
The Shannon Entropy of the entire grid can be computed as~\cite{bourgault2002information}:
\begin{equation}
    \mathbb{H}(\mathcal{OG}) = -\sum_{c_i \in \mathcal{OG}}p(c_i)log(p(c_i)) + (1-p(c_i))log(1-p(c_i))
    \label{entropy}
\end{equation}

To estimate the NBV, we compute the expected entropy-based information gain. As it is intractable to calculate the exact entropy from a predicted viewpoint, we perform a common simplifying approximation by predicting the expected measurements $\hat{z}^{view}_t$ from a viewpoint $a^{view}_t$ using ray-traversal algorithms. A sensor model representing our RGB-D sensor is defined with the given horizontal and vertical field of view (FoV) and resolution to cast a set of rays $\mathcal{R} = {r_1, r_2 , \dots r_j}$ for a given distance $d_{ray}$ in the \textit{z-axis} of the sensor model coordinate frame. A viewpoint $a^{view} \in \mathcal{A}^{view}$ is defined as the 3D position $\mathbf{p}^{view} \in \mathbb{R}^3$ and orientation $\mathbf{R}^{view} \in SO(3)$  of the camera frame. We perform Markov Monte-Carlo sampling of $N$ viewpoints on the hemisphere space located above the centroid $\mathbf{o}_{centroid}$ of the target object.
The size of the sphere is limited by the kinematic workspace limits of the robot.
The 3D position $\mathbf{p}^{view}$ is sampled as a point on the hemisphere and the orientation of the view as axis of rotation $\hat{\mathbf{e}}$ and angle $\theta$ is computed with
\begin{equation}
    \hat{\mathbf{h}} = \frac{\mathbf{p}^{view} - \mathbf{o}_{centroid}}{||\mathbf{p}^{view} - \mathbf{o}_{centroid} ||},
\end{equation}
\begin{equation}
    \theta = \cos^{-1}{(\hat{\mathbf{h}} \cdot \hat{\mathbf{Z}})}, \quad
    \hat{\mathbf{e}} = \frac{\hat{\mathbf{h}} \times \hat{\mathbf{Z}}}{|| \hat{\mathbf{h}} \times \hat{\mathbf{Z}}||},
    \label{eq:view_orientation}
\end{equation}
where $\hat{\mathbf{Z}} = \{0,0,1\}$ is the Z-axis of the world frame as shown in Figure~\ref{fig:actions}.
Using the resulting angle-axis formulation $(\hat{\mathbf{e}}, \theta)$ or equivalent rotation matrix $\mathbf{R}^{view}$ from \eqref{eq:view_orientation}, the camera is oriented towards the target object.
The grid cells which are traversed by the rays are computed to be occupied or free and the respective log-odds are updated accordingly~\cite{hornung2013octomap}:
\begin{equation}
    l(\hat{z}^{view}) = \left\{
                \begin{array}{ll}
                  log\frac{p_h}{1-p_h}  \quad \mathrm{if} \ \hat{z}^{view} \widehat{=} \textit{ hit} \\
                  log\frac{p_m}{1-p_m} \quad \mathrm{if} \ \hat{z}^{view}\widehat{=} \textit{ miss} 
                \end{array}
              \right.
    \label{eq:log-odds}
\end{equation}
where $p_h$ and $p_m$ are the probabilities of hit and miss which are user-defined values set to 0.7 and 0.4 respectively as in~\cite{hornung2013octomap}.
The expected information gain by taking a viewpoint $a^{view}_k$ and corresponding expected measurement $\hat{z}^{view}_t$ is given by the Kullback–Leibler (KL) divergence between the posterior entropy after integrating the expected measurements and the prior entropy~\cite{potthast2014probabilistic}:
\begin{equation}
    E[\mathbb{I}(p(c_i | a^{view}_t,  \hat{z}^{view}_t))] = \mathbb{H}(p(c_i)) - \mathbb{H}(p(c_i | a^{view}_t,  \hat{z}^{view}_t))
    \label{eq:kl_view}
\end{equation}
Hence, the selected action $a^{view*}$ is given by:
\begin{equation}
    a^{view*} = \argmax_{a_k^{view} \in \mathcal{A}^{view}}(E[\mathbb{I}(p(c_i | a^{view}_t,  \hat{z}^{view}_t))])
    \label{eq:kl_view_max}
\end{equation}

\subsubsection{Next Best Touch (NBT) Selection}\label{sec:nbt_select}
\begin{figure}[t!]
    \centering
    \includegraphics[width = \columnwidth, height = 5.3cm]{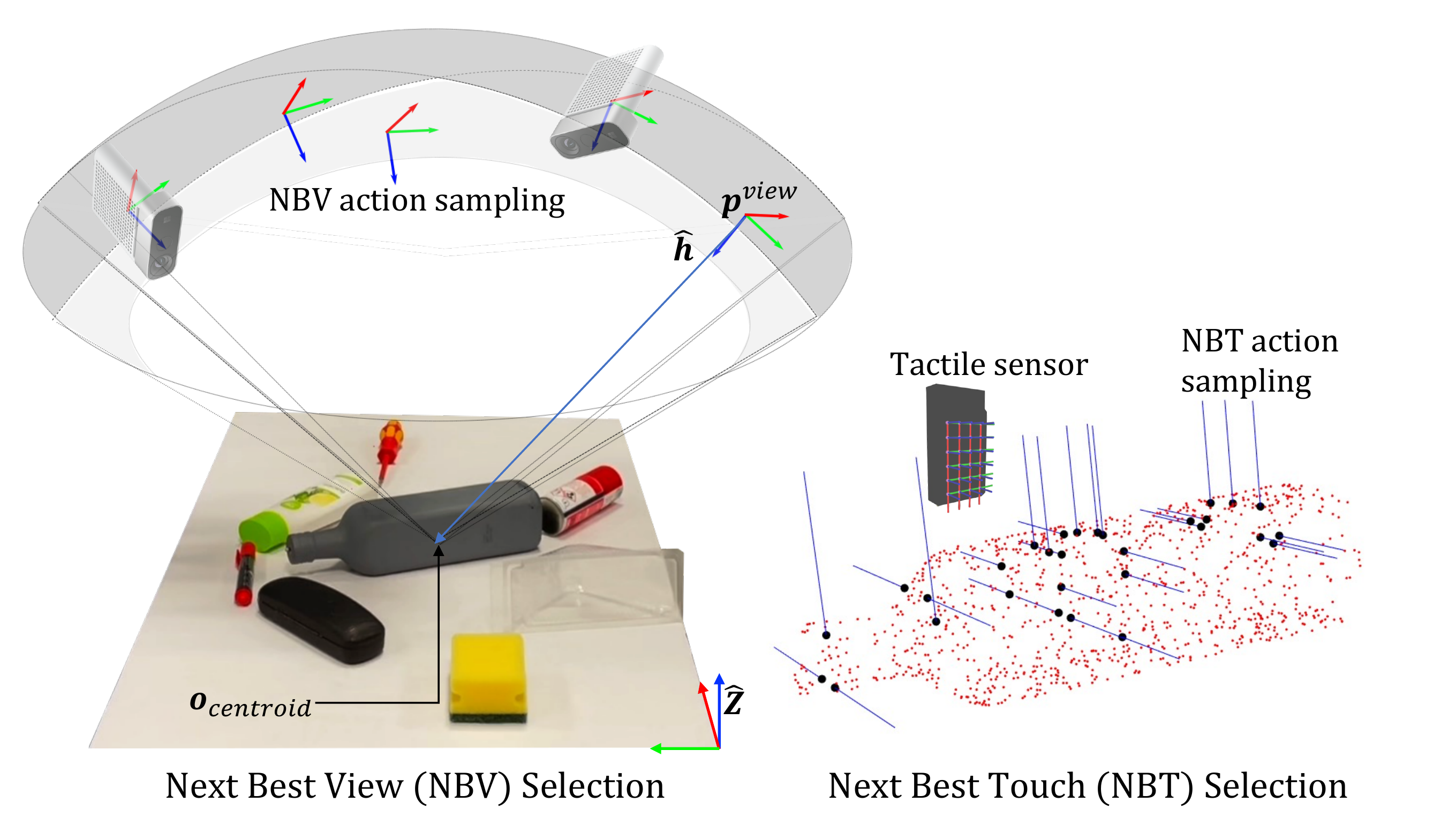}
    \caption{Monte-carlo sampling of visual viewpoints and touch points for next best view (NBV) and next best touch (NBT) action selection respectively.}
    \label{fig:actions}
\end{figure}

Similar to next best view selection, for tactile-based pose estimation we select the action to extract measurements that would reduce the uncertainty of the estimated pose.
We define an action $\mathbf{a}^{loc}_t$ as a ray represented by a tuple $\mathbf{a}^{loc}_t = (\mathbf{s}, \overrightarrow{\mathbf{d}})$, with $\mathbf{s}$ as the start point and $\overrightarrow{\mathbf{d}}$ the direction of the ray. 
The TIQF algorithm and active touch selection is initialised with minimum of 3 points, hence for initialisation the touches are sampled randomly given the visual-pose estimate.
We generate the set of possible actions $\mathcal{A}_{loc}$ by Monte-Carlo sampling of actions around each face of a bounding box placed on the current estimate of the object. The predicted measurement upon performing an action is estimated by ray-mesh intersection algorithm. We seek to choose the action $\mathbf{a}^{loc*}_{t} \in \mathcal{A}^{loc}$, that \textit{maximizes} the overall \textit{Information Gain} measured by the Kullback-Leibler divergence between the posterior distribution $p(\mathbf{x} | \hat{\mathbf{z}}_{1:t}, \hat{\mathbf{a}}_{1:t})$ after executing action $\hat{\mathbf{a}}_{t}$ and the prior distribution $p(\mathbf{x} | \mathbf{z}_{1:t-1}, \mathbf{a}_{1:t-1})$. We denote the predicted action and associated measurement as $\hat{\mathbf{z}}$ and $\hat{\mathbf{a}}$ respectively.
Given that the prior and posterior are multivariate Gaussian distributions from our definitions in the TIQF formulations, the KL divergence in discrete form can be computed in closed form as~\cite{duchi2007derivations}:
\begin{align}
    \mathbf{a}_{t}^{loc*} = \argmax_{\hat{a}_{t}} \frac{1}{2} &\left[ log\frac{det(\bar{\Sigma}_{t-1})}{det(\hat{\bar{\Sigma}}_{t})} + tr(\bar{\Sigma}_{t-1}^{-1} \hat{\bar{\Sigma}}_{t})) - d \right. \nonumber \\ 
    &\left. + (\hat{\bar{\mathbf{x}}}_{t} - \bar{\mathbf{x}}_{t-1})^T \bar{\Sigma}_{t}^{-1} (\hat{\bar{\mathbf{x}}}_{t} - \bar{\mathbf{x}}_{t-1}) \vphantom{log\frac{det(\Sigma_{t})}{det(\hat{\Sigma}_{t-1})}}\right],
    \label{eq:kld_closed_form}
\end{align}
where $d$ is the dimension of the quaternion state vector and $d=4$ in our case, $\bar{\mathbf{x}}_{t-1}$ is the normalized mean of the quaternion state estimate at $t-1$, and $\bar{\Sigma}^{\mathbf{x}}_{t-1}$ is the covariance matrix of the state at $t-1$. This enables to evaluate an exhaustive list of actions at marginal computation cost in \textit{real time} without the need to prune actions or setting trade-offs with computation time as compared to literature~\cite{tosi2014action, saund2017touch}. Timing analysis of our active action generation and selection approach is provided in our prior work~\cite{murali2021active}.
The next best action for pose estimation is graphically depicted in Figure~\ref{fig:actions}.

The stop criterion for both the NBV and NBT selection is defined similarly as the convergence criteria: if the change in position and rotation between each sensor acquisition is less than a specified threshold $\xi^{stop} = \{\xi^{stop}_T, \xi^{stop}_R \}$. In our experiments we set $\xi^{stop}_T = 5mm, \xi^{stop}_R = 2^{o}$. 

\section{Experiments}
\label{sec:experiments}
\subsection{Experimental Setup}
The experimental setup shown in Figure~\ref{fig:figure1} consists of a Universal Robots UR5 robot with a Robotiq 2F140 Gripper and a Franka Emika Panda robot with the standard Panda Gripper.
The standard gripper pads of the Robotiq 2F140 are replaced with the tactile sensor array from XELA Robotics on the fingertips and the phalanges. The tactile sensing system consists of $N_T = 140$ taxels that provide 3-axis force measurements on each taxel in the sensor coordinate frame. It is composed of eight tactile sensor arrays in total, where 4 tactile sensor arrays are on each finger: phalange sensor (24 taxels), outer finger (24 taxels), finger tip (6 taxels) and inner finger (16 taxels).
The tactile sensors function on the principle of Hall-effect sensing and are covered with a soft, textile material.
The raw data from the XELA sensor is a relative value of force measurement but it is not directly characterized to Newtons. The normal force values (along z axis) range between 36000 and 45000. We normalize the raw values received from the sensor.
An Azure Kinect DK RGB-D camera is rigidly attached to the Panda Gripper with a custom designed flange which provides the vision point cloud ${}^v\mathcal{S}$. Hand-eye calibration is performed to find the transformation between the Panda Gripper and the camera frame and consequently transformed into the common world coordinate frame $\mathcal{W}$~\cite{murali2021situ}. A marker-based optical tracking system from Advanced Realtime Tracking\footnote{https://www.ar-tracking.com/en} is placed overlooking the workspace which provides the ground-truth pose of the target objects only. The markers are placed only on the target object.

\begin{figure}[b!]
    \centering
    \includegraphics[width = 0.8\columnwidth, height=3.5cm]{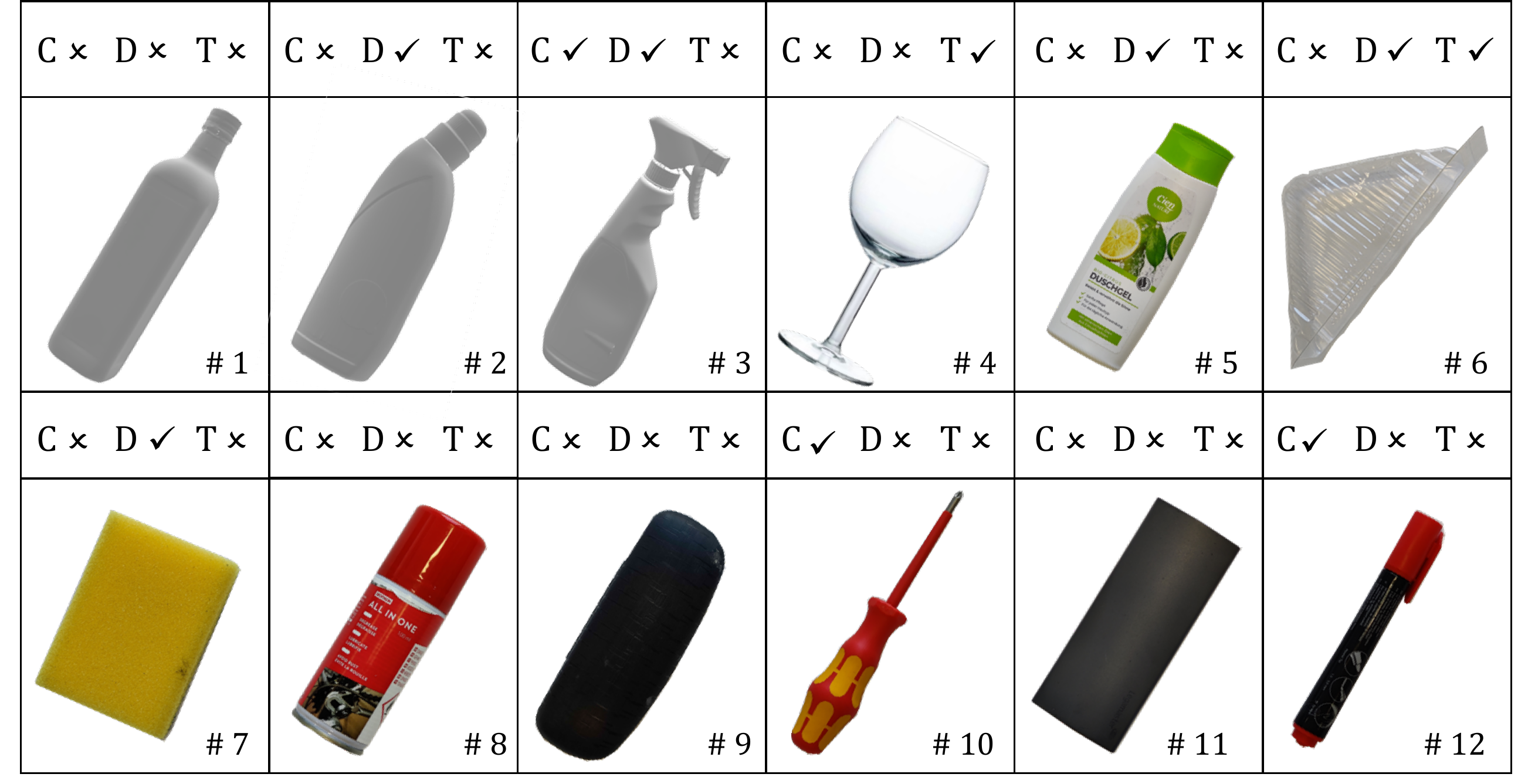}
    \caption{Experimental objects. The properties evaluated by human subjects: C: complex shape, D: deformability, T: transparency. $\checkmark$: high, $\times$: low. }
    \label{fig:obj_list}
\end{figure}

We used 12 objects in total: olive oil bottle, cleaner, spray, transparent wineglass, shampoo, transparent box, sponge, can, black box, screwdriver, duster, marker as shown in Figure~\ref{fig:obj_list}. 
The objects have been chosen according to the following criteria: varying shape between simple (e.g. can) to complex (e.g. screwdriver), varying degrees of transparency (highly transparent box to highly opaque black box), varying center of mass (e.g. shampoo) and varying degree of deformability (e.g. sponge). 
Some of the objects such as the transparent box and wineglass are intentionally chosen to test the robustness of the framework. The background has been intentionally chosen to be plain white to increase the visual perceptual difficulty of the transparent objects. 
Four objects i.e., olive oil bottle, spray, cleaner and transparent wine glass are used as the target object whose pose needs to be accurately estimated while the other 8 objects are used to clutter the workspace. A software architecture developed in ROS is used for the data communication between the two robots, camera, and tactile sensors.
For the implementation of the finite state machine, we used the Octomap library~\cite{hornung2013octomap} for the NBV calculations.

The robot experiments were executed on a workstation running Ubuntu 18.04 with 8 core Intel i7-8550U CPU @ 1.80GHz and 16 GB RAM. 
The maximum allowed speeds for the UR5 and Panda were 75 mm/s and 100 mm/s respectively for safety constraints.
The fine tuning of the semantic segmentation network~\cite{chen2017deeplab} employed NVidia GeForce RTX 2080 Super GPU with 8GB RAM. 
No further training of the grasp affordance network~\cite{morrison2020learning} was performed.

\begin{figure}[t!]
 \centering
\includegraphics[width=\columnwidth, height = 4.5cm]{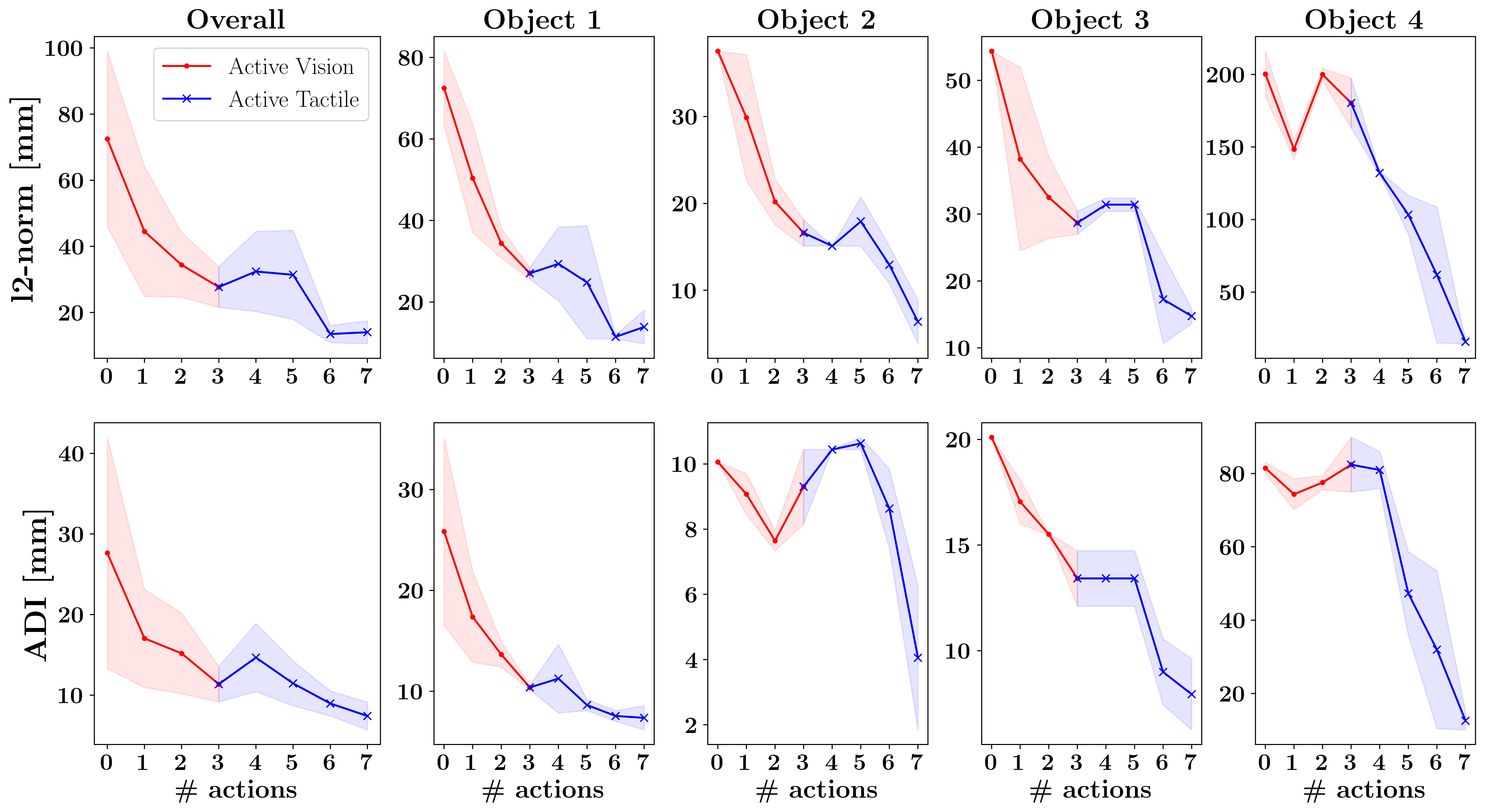}
\caption{Plots showing the mean and standard deviation for $L_2$ norm error (translation) (top) and ADI (bottom) for the four target objects after decluttering. The overall plot shows the median and median absolute deviation of the combined data from all objects.}
\label{fig:std_mean_plot}
\end{figure}


\begin{table*}[th!]
\centering
\caption{Ablation studies for baselines (a) static vision, active vision (b) before and (c) after decluttering and our proposed active visuo-tactile with decluttering method.}
\label{tab:results}
\resizebox{\textwidth}{!}{%
\begin{tabular}{@{}lllllllll@{}}
\toprule
 & \multicolumn{2}{c}{\textbf{Static vision}} & \multicolumn{2}{c}{\textbf{\begin{tabular}[c]{@{}c@{}}Active vision without\\  decluttering\end{tabular}}} & \multicolumn{2}{c}{\textbf{\begin{tabular}[c]{@{}c@{}}Active vision with\\  decluttering\end{tabular}}} & \multicolumn{2}{c}{\textbf{\begin{tabular}[c]{@{}c@{}}Active visuo-tactile with\\  decluttering\end{tabular}}} \\
 & \multicolumn{1}{c}{ADI (mm) $\downarrow$} & \multicolumn{1}{c}{\begin{tabular}[c]{@{}c@{}}$L_2$ norm (Trans)\\ (mm) $\downarrow$\end{tabular}} & \multicolumn{1}{c}{ADI (mm) $\downarrow$} & \multicolumn{1}{c}{\begin{tabular}[c]{@{}c@{}}$L_2$ norm (Trans)\\ (mm) $\downarrow$\end{tabular}} & \multicolumn{1}{c}{ADI (mm) $\downarrow$} & \multicolumn{1}{c}{\begin{tabular}[c]{@{}c@{}}$L_2$ norm (Trans)\\ (mm) $\downarrow$\end{tabular}} & \multicolumn{1}{c}{ADI (mm) $\downarrow$} & \multicolumn{1}{c}{\begin{tabular}[c]{@{}c@{}}$L_2$ norm (Trans)\\ (mm)$\downarrow$\end{tabular}} \\ \midrule
Obj 1 & 41.90 $\pm$ 4.56 & 118.44 $\pm$ 38.35 & 11.54 $\pm$ 1.88 & 40.61 $\pm$ 1.18 & 10.39 $\pm$ 0.34 & 27.04 $\pm$ 2.01 & \textbf{7.38} $\pm$ 1.68 & \textbf{13.88} $\pm$ 5.81 \\
Obj 2 & 34.60 $\pm$ 1.54 & 159.72 $\pm$ 62.29 & 14.71 $\pm$ 0.76 & 152.59 $\pm$ 151.73 & 9.31 $\pm$ 1.61 & 16.61 $\pm$ 2.13 & \textbf{4.06} $\pm$ 3.12 & \textbf{6.38} $\pm$ 3.58 \\
Obj 3 & 49.79 $\pm$ 3.99 & 169.76 $\pm$ 7.91 & 35.99 $\pm$ 1.14 & 104.82 $\pm$ 40.19 & 13.42 $\pm$ 1.86 & 31.42 $\pm$ 1.40 & \textbf{7.94} $\pm$ 2.38 & \textbf{14.75} $\pm$ 1.58 \\
Obj 4 & 216.22 $\pm$ 100.48 & 368.07 $\pm$ 58.98 & 63.88 $\pm$ 29.32 & 224.38 $\pm$ 20.95 & 82.4 $\pm$ 10.56 & 180.29 $\pm$ 24.28 & \textbf{12.57} $\pm$ 3.52 & \textbf{15.75} $\pm$ 1.83 \\
Mean & 85.63 $\pm$ 87.28 & 203.99 $\pm$ 111.61 & 31.53 $\pm$ 24.15 & 130.60 $\pm$ 77.55 & 28.88 $\pm$ 35.72 & 63.84 $\pm$ 77.88 & \textbf{7.99} $\pm$ 3.50 & \textbf{12.69} $\pm$ 4.28 \\
Median & 45.85 $\pm$ 7.59 & 164.74 $\pm$ 25.66 & 25.35 $\pm$ 12.23 & 128.71 $\pm$ 55.99 & 11.90 $\pm$ 2.06 & 29.23 $\pm$ 7.40 & \textbf{7.66} $\pm$ 1.94 & \textbf{14.31} $\pm$ 0.94 \\ \bottomrule
\end{tabular}%
}
\end{table*}

\subsection{Robot Experiment Results}
Given the estimated pose $\mathbf{R}_{est}, \mathbf{t}_{est}$ and the ground truth poses $\mathbf{R}_{gt}, \mathbf{t}_{gt}$, we employ the model-free translation and rotation error metric and the model-dependent Average Distance of model points with Indistinguishable views metric (ADI)~\cite{hinterstoisser2012model} for evaluation. The translation and rotation error is defined as follows: 
\begin{align}
    \mathtt{err_T} & = ||\mathbf{t}_{est} -\mathbf{t}_{gt}||_2 , \\
    \mathtt{err_R} & = \cos^{-1}((Tr(\mathbf{R}_{est}\mathbf{R}_{gt}^{-1})-1)/2)
\end{align}
where, $||x||_2$ is the $L_2$ norm of $x$. As objects having an axis of symmetry can produce infinite rotational solutions, we only report the $L_2$ norm of translation error for all the objects. Instead, we use the ADI metric as it is not affected by symmetric objects. The ADI metric is defined as follows: 
\begin{equation}
    \mathtt{err}_{adi} = \frac{1}{M}\sum_{\mathbf{p}_1 \in \mathcal{O}} \min_{\mathbf{p}_2 \in \mathcal{O}} || (\mathbf{R}_{gt}\mathbf{p}_1 + \mathbf{t}_{gt}) - (\mathbf{R}_{est}\mathbf{p}_2 + \mathbf{t}_{est}) ||,
    \label{eq:adi}
\end{equation}
for all points $p_1, p_2 \in \mathcal{0}$ and $M$ is the total number of points in $\mathcal{O}$.
For both the metrics, lower values signify higher accuracy.

Considering the implementation of robot actions, we used both vision and tactile feedback for the push, grasp and touch actions. For instance, given a push or grasp action, the tactile readings are continuously sampled at 40 Hz to detect possible loss of contact during pushing or grasping.  We use a constant grasping force of 5N provided by the Robotiq 2F140 gripper. For the push actions, the contacted taxels' normalized raw force values are monitored such that they are constantly above a predefined threshold i.e., $f_r > \tau_p$ (set to 1.06).
On the other hand, for touch actions for localization, we use guarded motions so that the robot does not accidentally push or topple other objects or the target object. As soon as the normalized force value measured on any of the taxels exceeds the threshold $f_r > \tau_f$ (set to 1.02), the motion is stopped and the 3D locations of the excited taxels are recorded as the tactile point cloud ${}^{t}\mathcal{S}$. 
We compare our active visuo-tactile pose estimation by decluttering the scene with 3 baselines: (a) static vision without decluttering, (b) active vision without decluttering and (c) active vision with decluttering. This ablation study is performed to evaluate the importance of each part of the framework. In all cases, the pose estimation is performed using our TIQF algorithm with the same initial conditions and scene segmentation to ensure uniformity.
We repeated all the baseline experiments and our proposed framework twice for each target object by randomly changing the scene clutter each time. In total, we performed 32 experimental trials including baselines.
The results for the experiments are shown in Table~\ref{tab:results}. Figure~\ref{fig:std_mean_plot} shows the accuracy of the pose estimation using $L_2$ norm of the translation error and ADI for the \textit{active} vision and \textit{active} touch-based pose estimation. 
A typical run of the whole framework consisting of 4 objects to declutter, followed with 3 different viewpoints for active vision and 4 touch-acquisitions respectively takes around 795s, while 87\% of the time is used for robot actions alone.
We also report an overall success rate of $83.3 \%$ for grasp actions and $70\%$ for push actions for the decluttering phase.

\subsection{Discussion}
As seen from Table~\ref{tab:results}, the ADI metric and the $L_2$ norm decreases and accuracy improves from static vision to our proposed active visuo-tactile estimation with decluttering approach. 
We note approximately $44.7\%$ reduction in median ADI error with active vision compared to static vision. This corroborates with prior work~\cite{wu2015active}, wherein selecting viewpoints actively can improve accuracy over static viewpoints. 
Moreover, demonstrating the validity of our proposed decluttering strategy, we see a reduction of $53 \%$ in median ADI error before and after decluttering.
On the other hand, active vision-based pose estimation on a scene without clutter may still have residual uncertainty. This is demonstrated by the improved performance of $35.6\%$ in median ADI using active tactile-based pose estimation. 
We intentionally used a transparent wineglass as a target object which is very challenging for pose estimation from visual perception as it is nearly invisible to a time-of-flight (ToF) depth sensor. This is seen by the relatively higher errors in Table~\ref{tab:results} (Object \#4) in comparison to other target objects. However, visuo-tactile based estimation using TIQF reduces the ADI error by nearly 85\% compared to active vision after decluttering. The errors are consistent with other target objects, highlighting the strength of tactile sensing for challenging objects for visual modality.
Furthermore, the ability of the TIQF to handle dense and sparse clouds is shown by the improved accuracy in vision and tactile-based estimation respectively over each action as shown in Figure~\ref{fig:std_mean_plot}. 
The TIQF converges to a stable pose estimate with $<1cm$ average error within 4 touches.
In Figure~\ref{fig:std_mean_plot}, we note that a change in modality from active vision to active tactile during interactive perception helps to improve the accuracy of pose estimation. 

\section{Conclusions}
\label{sec:conclusion}
In this paper, we proposed an active visuo-tactile pose estimation framework for objects in dense clutter.
We proposed a novel declutter graph based approach for scene representation for decluttering which allows to select the next object to remove and provides the optimal action to perform. The declutter scene graph further encodes two types of actions: push and grasp action. Furthermore, we extended our novel TIQF for active vision based and active tactile based pose estimation. We performed an object-driven exploration strategy for active viewpoint and active touch point selection. In the evaluation, we demonstrated that our proposed method significantly improves the accuracy of pose estimation over mono-modal baselines. 
We also demonstrated the importance of using a secondary modality to correct or verify the estimation from a first modality.

\section*{Acknowledgment}
The video can be found in: \url{https://youtu.be/sjqWRFLL2Xw}

\bibliography{biblio}
\bibliographystyle{IEEEtran}
\end{document}